\documentclass[letterpaper, 10 pt, conference]{ieeeconf}  
\IEEEoverridecommandlockouts                              
\usepackage{amsmath,amsfonts}
\usepackage{algorithmic}
\usepackage{algorithm}
\usepackage{array}
\usepackage{textcomp}
\usepackage{stfloats}
\usepackage{url}
\usepackage{verbatim}
\usepackage{graphicx}
\usepackage{cite}
\usepackage[dvipsnames]{xcolor}

\usepackage{amssymb}
\usepackage{rotating}
\usepackage{color}
\usepackage{framed}
\usepackage{multirow}
\usepackage{subcaption}
\usepackage{balance}
\usepackage{booktabs}
\usepackage{url}
\usepackage{bm}
\usepackage{bbm}
\usepackage{arydshln}
\usepackage{xspace}
\usepackage[export]{adjustbox}
\usepackage{lipsum}

\usepackage{framed,multirow}
\usepackage{colortbl}
\usepackage{latexsym}
\usepackage{graphics}
\usepackage{duckuments}
\usepackage{tablefootnote}
\usepackage{mathptmx}
\usepackage{gensymb}
\usepackage{makecell}
\usepackage{dsfont}

\def \ie {\emph{i.e.}}
\def \eg {\emph{e.g.}}
\def \etal {\emph{et al.}}
\newcommand{\tit}[1]{\smallbreak\noindent\textbf{#1.}}

\newcommand{\tinytit}[1]{\noindent\textbf{#1.}}

\newcommand{\mbb}[1]{\mathbf{#1}}

\title{\LARGE \bf Embodied Agents for Efficient Exploration and Smart Scene Description
}

\author{Roberto Bigazzi, Marcella Cornia, Silvia Cascianelli, Lorenzo Baraldi, Rita Cucchiara
\thanks{R. Bigazzi, M. Cornia, S. Cascianelli, L. Baraldi, R. Cucchiara are with the University of Modena and Reggio Emilia, Italy 
    {\tt\small \{firstname.lastname\}@unimore.it}}%
}

\begin{document}

\maketitle
\thispagestyle{empty}
\pagestyle{empty}

\begin{abstract}
The development of embodied agents that can communicate with humans in natural language has gained increasing interest over the last years, as it facilitates the diffusion of robotic platforms in human-populated environments. As a step towards this objective, in this work, we tackle a setting for visual navigation in which an autonomous agent needs to explore and map an unseen indoor environment while portraying interesting scenes with natural language descriptions. To this end, we propose and evaluate an approach that combines recent advances in visual robotic exploration and image captioning on images generated through agent-environment interaction. Our approach can generate smart scene descriptions that maximize semantic knowledge of the environment and avoid repetitions. Further, such descriptions offer user-understandable insights into the robot’s representation of the environment by highlighting the prominent objects and the correlation between them as encountered during the exploration. To quantitatively assess the performance of the proposed approach, we also devise a specific score that takes into account both exploration and description skills. The experiments carried out on both photorealistic simulated environments and real-world ones demonstrate that our approach can effectively describe the robot's point of view during exploration, improving the human-friendly interpretability of its observations.
\end{abstract}


\section{Introduction}

Over the past few years, advances on visual navigation and machine learning have shown that training navigation policies with reinforcement learning in simulated photorealistic environments~\cite{xia2018gibson,chang2017matterport3d,savva2019habitat} for millions or even billions of frames~\cite{wijmans2019dd} allows agents to generalize their navigation skills to unseen environments without the need of acquiring any previous knowledge. A plethora of exploration rewards and strategies has been presented in the representation learning community and, generally, those strategies can be adapted to work on board of smart autonomous agents and improve their perception of the surrounding, and consequently, the desired behavior. A popular approach for such agents follows a modular, hierarchical navigation policy design such as the one proposed by Chaplot~\etal~\cite{chaplot2019learning}, which optimizes a reward function that encourages exploration~\cite{ramakrishnan2020occupancy,ramakrishnan2021exploration,bigazzi2022impact}. Embodied agents trained on sufficiently photorealistic simulators are then deployable onto real-world robotic platforms~\cite{kadian2020sim2real,bigazzi2021out,truong2021bi}, thus contributing to the widespread of service robots.

At the same time, research connecting robotics, vision, and natural language processing has attracted interest and has led to the emergence of robotic vision and language tasks, such as vision-and-language navigation~\cite{anderson2018vision,landi2019perceive,krantz2020beyond,anderson2021sim} and embodied question answering~\cite{das2018embodied,das2018neural,wijmans2019embodied,yu2019multi}. 
However, using language to facilitate the human understanding of the behavior and perception of a robotic agent is still an under-explored path~\cite{cascianelli2018full,cornia2019smart,bigazzi2020explore,tan2022embodied}, in which further research steps are needed to reach a seamless interaction between humans and robots. 
In this respect, image captioning approaches, whose goal is to generate a natural language description of a given image, have seldom been employed in navigation and exploration settings. In such approaches, images can be represented by using convolutional neural networks to extract global features~\cite{karpathy2015deep}, grids of features~\cite{xu2015show}, or features for image regions containing visual entities~\cite{anderson2018bottom}. 
In most cases, attention mechanisms are applied to enhance the visual input representation. More recent approaches employ fully-attentive Transformer-like architectures~\cite{vaswani2017attention} as visual encoders, which can also be applied directly to image patches~\cite{liu2021cptr,cornia2021universal}.
The image representation is then used to condition a recurrent neural network~\cite{karpathy2015deep,rennie2017self,huang2019attention} or a Transformer-based~\cite{cornia2020meshed,zhang2021rstnet} language model that generates the caption. 

The capability of image captioning approaches to produce natural descriptions of everyday images is, however, insufficient when working in an embodied setting in which the agent continuously moves inside of an environment and needs to portray interesting scenes. In such a scenario, indeed, not all the images observed by the agent are semantically relevant as they might contain uninteresting objects or scenes (\eg~when facing a wall) or there might be low variety between temporally consecutive observations. Hence, a naive captioning of such images would result in uninteresting descriptions and avoidable repetitions. To overcome this, the captions should be generated only when the scene observed is worth being described to a human.

In this work, we take a step forward with respect to the above-mentioned limitations and propose a complete pipeline for efficient exploration and mapping which can, at the same time, provide user-understandable representations of the perceived environment in the form of natural language descriptions.
We jointly integrate and propose visual exploration strategies, a state-of-the-art approach for image captioning and smart description policies in an embodied setting, with the final aim of improving human understanding of robotic perception. Also, we devise a novel metric, called episode description score (\textbf{$\mathsf{ED}\text{-}\mathsf{S}$}), that evaluates the exploration and the ability of covering objects in the environment avoiding repetitions. We extensively test the performance of the proposed approach in comparison with different baselines on both Gibson~\cite{xia2018gibson} and Matterport3D~\cite{chang2017matterport3d} datasets.
Finally, while our approach is trained and evaluated in simulation, the proposed architecture is designed for the final deployment on a real robotic platform, as we show in the video accompanying the submission.
\section{Proposed Approach}

\begin{figure*}
    \centering
    \includegraphics[width=0.85\textwidth, height=6cm]{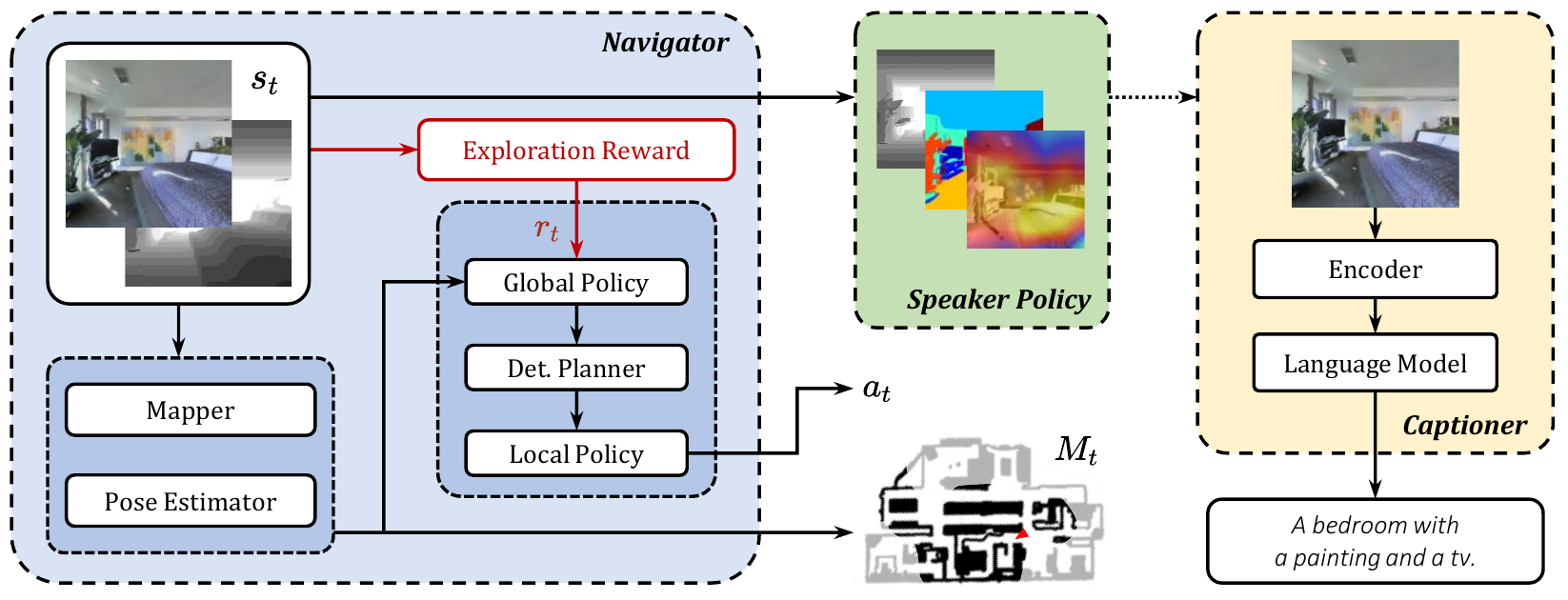}
    \caption{Overview of the proposed approach for smart scene description, comprising a navigator, a speaker policy, and a captioner module.}
    \vspace{-0.5cm}
    \label{fig:ex2_system}
\end{figure*}

Our proposed architecture is composed of three main components: a \textit{navigator}, in charge of the exploration, a \textit{captioner}, that describes interesting scenes, and the \textit{speaker policy} that decides when the captioner should be activated. An overview of our complete architecture is shown in Fig.~\ref{fig:ex2_system}.

\subsection{Navigator}
\label{sec:navigation_method}
The exploration capabilities of the agent are strictly dependant on the performance of the navigation module, therefore relying on a proper navigation approach is of fundamental importance. Following recent literature on embodied visual navigation \cite{chaplot2019learning,ramakrishnan2020occupancy,ramakrishnan2022poni}, we devise a hierarchical policy coupled with a learned neural occupancy mapper and a pose estimator.
The hierarchical policy sets long and short-term navigation goals, while the neural mapper builds an occupancy grid map representation of the environment and the pose estimator locates the agent on such map.
 
\tit{Mapper}
In order to track explored and unexplored regions of the environment over time, the mapper is a fundamental component. Moreover, a neural-based mapper allows to infer regions occupancy beyond the observable area in front of the agent, facilitating the planning phase~\cite{ramakrishnan2020occupancy}. The output of the mapper is a $M \times M \times 2$ global map of the environment $m_t$ that keeps track of the non-traversable space in its first channel and the area explored by the agent in the second one.
At each time step, the mapper processes the RGB-D observation $s_t=(s^{rgb}_t, s^{d}_t)$ coming from the agent and predicts a $L \times L \times 2$ egocentric local map $l_t$ representing the state in front of the agent. RGB images are encoded using a ResNet-18~\cite{he2016deep} followed by a UNet encoder~\cite{ronneberger2015u}, while depth observations are encoded using only a UNet encoder. RGB and depth features are concatenated and fed to a CNN to merge the two modalities. Finally, a UNet decoder processes the merged features to predict the local map $l_t$. At every timestep $t$, the local map $l_t$ is transformed using the estimated pose of the agent $\omega_t$ and registered to the global map $m_t$ with a moving average. The global map is initially empty and is built incrementally with the exploration of the environment. 

\tit{Pose Estimator}
Relying on a global map requires a robust pose estimator in order to build geometrically coherent and precise maps. Indeed, an inaccurate pose estimate would rapidly diverge from the ground-truth pose, and loop closure is inapplicable if previous knowledge of the environment is not available. Furthermore, directly using the pose sensor of the robot is not sufficient since sensor noise, slipping wheels, and collisions with obstacles would not be accounted for. 
The adopted approach uses the difference between consecutive pose sensor readings $\Delta \omega'_{t}=\omega'_{t}-\omega'_{t-1}$ as a first estimate of the motion of the agent, where $\omega'_t=(x'_t, y'_t, \theta'_t)$, with $(x'_t,y'_t)$ being the coordinates on the map, and $\theta'_t$ the orientation of the agent. In order to correct eventual inaccuracies, we use local maps $l_{t}$, $l_{t-1}$ extracted from the respective observations as feedback. The local map $l_{t-1}$ is rototranslated with respect to the current position of the agent using $\Delta \omega'_{t}$. Transformed $l_{t-1}$ and $l_t$ are concatenated and fed to a CNN to output a corrected displacement $\Delta \omega_{t}$. At every timestep, $\Delta \omega_{t}$ is used to compute the pose of the agent with respect to the pose at the previous step:
\begin{equation}
    \omega_{t} = \omega_{t-1} + \Delta \omega_{t} \quad \text{where} \quad \omega_t = (x_t, y_t, \theta_t).
\label{eq:pose_estimator}
\end{equation}
Without loss of generality, we consider the agent starting from $\omega_0=(0,0,0)$, \ie~the center of the global map $m_t$.

\tit{Navigation Policy}
The navigation policy adopts a hierarchical structure as used in~\cite{chaplot2019learning,ramakrishnan2020occupancy,ramakrishnan2022poni}. Specifically, the navigation policy comprehends three modules: a high-level global policy, a deterministic planner, and an atomic local policy. 
The hierarchical policy is adopted to decouple high-level and low-level concepts like moving across rooms and avoiding obstacles. It samples a goal coordinate on the map, while the deterministic planner uses the global goal to compute a local goal in close proximity of the agent. The local policy then predicts actions to reach the local goal.

\begin{figure*}
    \centering
    \footnotesize
    \resizebox{\linewidth}{!}{
        \setlength{\tabcolsep}{.15em}
        \begin{tabular}{ccccc}
             \textbf{Curiosity} & 
             \textbf{Coverage} & 
             \textbf{Anticipation} & 
             \textbf{Impact (Grid)} & 
             \textbf{Impact (DME)} \\
            \includegraphics[width=0.20\textwidth]{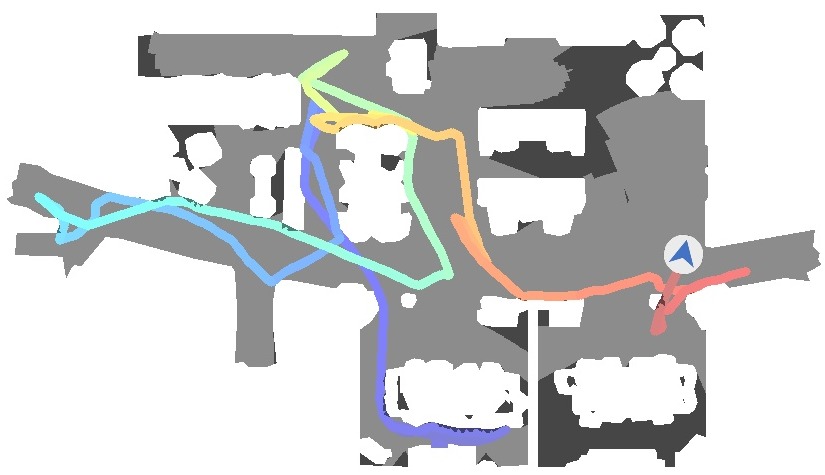} &
            \includegraphics[width=0.20\textwidth]{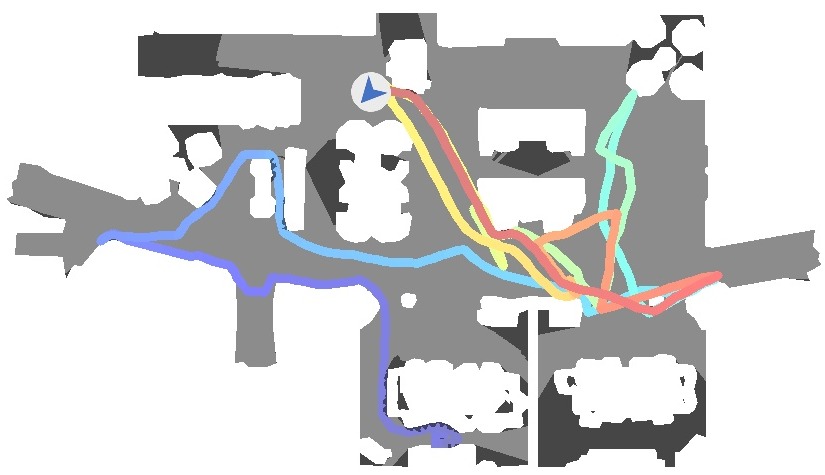} &
            \includegraphics[width=0.20\textwidth]{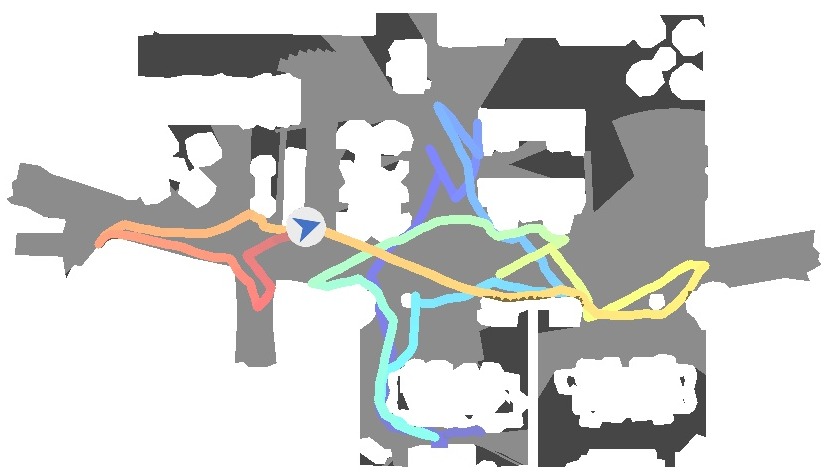} &
            \includegraphics[width=0.20\textwidth]{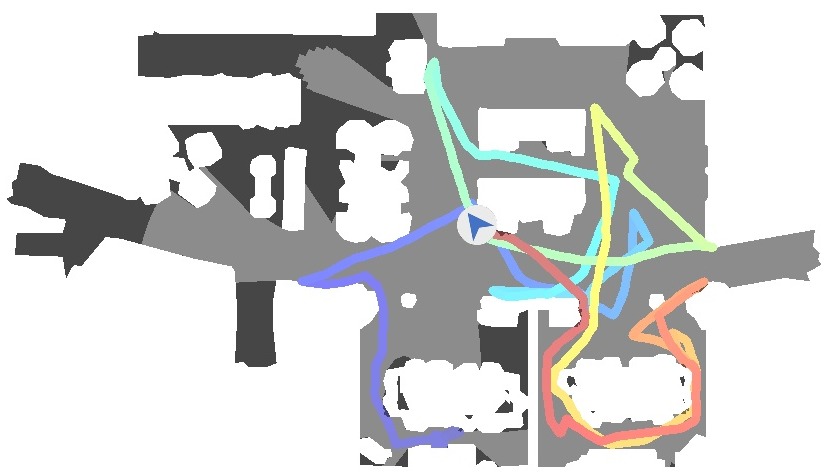} &
            \includegraphics[width=0.20\textwidth]{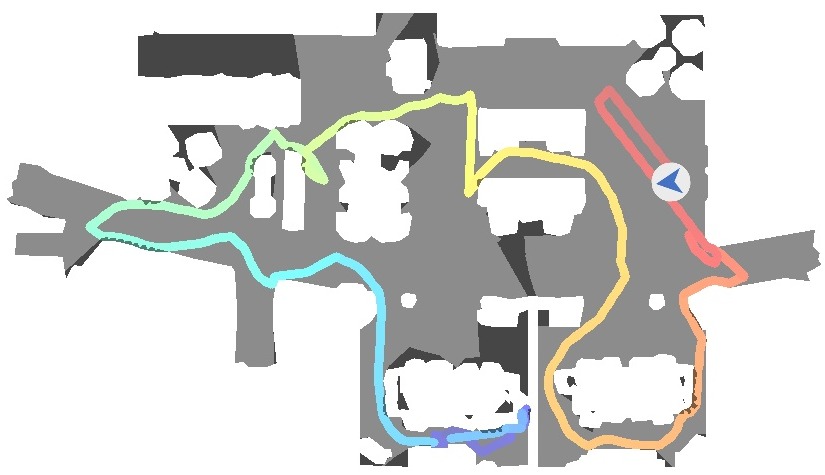}
        \end{tabular}
    }
    \caption{Qualitative exploration trajectories of different navigation agents on the same episode.}
    \vspace{-0.5cm}
    \label{fig:navigation_mp3d_maps}
\end{figure*}

The global policy takes as input an enriched version of the current global map $m_t$. A map encoding of the current position of the agent and a map with already visited locations are concatenated to $m_t$ to form a $M \times M \times 4$ map $m^+_t$. The enriched map $m^+_t$ is both max-pooled and cropped with respect to the position of the agent to a dimension $G \times G \times 4$. Then, the resulting two versions of $m^+_t$ are stacked together to form the final 8-channel input of the global policy which is processed to sample a point on a $G \times G$ grid. The output of the global policy is converted to a coordinate on the global map $m_t$, that is the global goal $g_t$. The global policy is trained with reinforcement learning using the global reward $r^{global}_t$.

The deterministic planner adopts the A* algorithm to compute a feasible trajectory from the current position of the agent to the global goal using the current state of the map $m_t$. A point on the trajectory within $1.25$m from the agent is extracted to form the local goal $p_t$.

The local policy takes as input the current RGB observation $s^{rgb}_t$ as well as the relative displacement of the local goal $p_t$ from agent's position $\omega_t$, and predicts the atomic action needed to reach the local goal. The output of the local policy corresponds to one of the following atomic actions: \textit{move forward 0.25m}, \textit{turn left 10\textdegree}, and \textit{turn right 10\textdegree}. This policy is trained with a reward $r^{local}_t$ that encourages the decrease in the geodesic distance between agent and local goal:
\begin{equation}
    r^{local}_{t} = d(m_{t-1},p_{t-1},\omega_{t-1}) - d(m_{t},p_{t},\omega_{t}),
\label{eq:local_reward}
\end{equation}
where $d(\cdot)$ returns the geodesic distance using the position of the agent, the local goal, and the global map to account for possible obstacles on the way.

\subsection{Exploration Rewards}
We compare various global exploration rewards such as curiosity~\cite{pathak2017curiosity}, coverage~\cite{chaplot2020object}, anticipation~\cite{ramakrishnan2020occupancy}, and impact~\cite{bigazzi2022impact}. All the considered methods obtain the reward by exploiting visual input sensors only.
Exemplar exploration trajectories resulting from the different rewards are reported in Fig.~\ref{fig:navigation_mp3d_maps}.

\tit{Curiosity}
The curiosity reward adopts two additional neural networks that learn the environment dynamics. A forward model is trained to predict the encoding of the future RGB observation given the encoding of the current observation and action, and an inverse model is trained to infer the action $a_t$ performed between consecutive observations $(s^{rgb}_{t},s^{rgb}_{t+1})$. These models are trained minimizing the following losses:
\begin{equation}
    L_{fwd} = \frac{1}{2} \left\| \hat\phi(s^{rgb}_{t+1}) - \phi(s^{rgb}_{t+1}) \right\|^2_2  \quad\mathrm{and}\quad 
    L_{inv} = {y_t \log \hat a_t},
\label{eq:curiosity_losses}
\end{equation}
where $\hat\phi(s^{rgb})$ and $\phi(s^{rgb})$ denote predicted and ground-truth encodings of the observation $s^{rgb}$, $y$ is the one-hot encoding of the ground-truth action $a$, and $\hat{a}$ is the predicted action probability distribution.
The global reward for the curiosity-driven model is given by the error of the forward dynamics model prediction during the navigation:
\begin{equation}
        r^{global}_t = \frac{\eta}{2} \| \hat\phi(s^{rgb}_{t+1}) - \phi(s^{rgb}_{t+1}) \|^2_2,
\label{eq:curiosity_reward}
\end{equation}
where $\eta$ is a normalizing term set to $0.01$.

\tit{Coverage}
The coverage-based reward maximizes the information gathered at each timestep, being it the number of objects or landmarks reached or area seen. In this work, we consider the area seen definition, as proposed in~\cite{chaplot2019learning}:
\begin{equation}
    r^{global}_t = \text{AS}_{t} - \text{AS}_{t-1},
\label{eq:coverage_reward}
\end{equation}
where AS indicates the number of pixels explored in the ground-truth map.

\tit{Anticipation}
The occupancy anticipation reward~\cite{ramakrishnan2020occupancy} aims to maximize accuracy in the prediction of the map including occluded unseen areas, \ie,
\begin{align}
    r^{global}_t = \text{Acc}(m_t, m^\star) - \text{Acc}(m_{t-1}, m^\star), \\
    \text{Acc}(m, m^\star) = \sum_{i=1}^{M^2}\sum_{j=1}^2\mathds{1}[m_{ij} = m^\star_{ij}],
\label{eq:occant_reward}
\end{align}
where $m$ is the predicted global map, $m^\star$ is the ground-truth global map, and $\mathds{1}[\cdot]$ is the indicator function.

\tit{Impact}
The impact reward encourages actions that modify agent's internal representation of the environment, with impact at timestep $t$ that is measured as the $l_2$-norm of the encodings of the two consecutive states $\phi(s_{t})$ and $\phi(s_{t+1})$. However, using the formulation of impact as it is, could lead to trajectory cycles with high impact but low exploration. To overcome this issue, Raileanu~\etal~\cite{raileanu2020ride} uses the state visitation count $N(s_t)$ to scale the reward. Unfortunately in our setting, the concept of the visitation count is not directly applicable, due to the continuous space of the photo-realistic environment. Hence, we adopt and evaluate the impact-based methods proposed in~\cite{bigazzi2022impact}. Such methods formalize a pseudo-visitation count $\widehat{N}(s_t)$ in continuous environments with two different approaches: grid and density model estimation. The final global reward for the impact-driven model becomes:
\begin{equation}
    r^{global}_t = \left\| \phi(s^{rgb}_{t+1}) - \phi(s^{rgb}_{t}) \right\|_2\Big/\sqrt{\widehat{N}(s_t)},
\label{eq:impact_reward_dme}
\end{equation}
where $\phi(s^{rgb})$ and $\widehat{N}(s_t)$ are the encoding and the estimated pseudo-visitation count at timestep $t$.

\subsection{Captioner}\label{ssec:cap_module}
The goal of the captioning module is that of modeling an autoregressive distribution probability $p(\bm{w}_t|\bm{w}_{\tau<t}, \mbb{V})$, where $\mbb{V}$ is an image captured from the agent and $\{\bm{w}_t\}_t$ is the sequence of words comprising the generated caption. This is usually achieved by training a language model conditioned on visual features to mimic ground-truth descriptions. 
For multimodal fusion, we employ an encoder-decoder Transformer~\cite{vaswani2017attention} architecture. Each layer of the encoder employs multi-head self-attention (MSA) and feed-forward layers, while each layer of the decoder employs multi-head self- and cross-attention (MSCA) and feed-forward layers. For enabling text generation, sequence-to-sequence attention masks are employed in each self-attention layer of the decoder. 

To obtain the set of visual features $\mbb{V}$ for an image, our model employs a visual encoder that is pre-trained to match vision and language (\ie~CLIP~\cite{radford2021learning}). Compared to using features extracted from object detectors~\cite{anderson2018bottom,zhang2021vinvl}, 
our strategy is beneficial in terms of both computational efficiency and feature quality. 
The visual descriptors $\mbb{V}=\left[ \bm{v}_1, \bm{v}_2, ..., \bm{v}_N \right]$ are encoded via bi-directional attention in the encoder, while the token embeddings of the caption $\mbb{W} = \left[ \bm{w}_1, \bm{w}_2, ..., \bm{w}_L\right]$ are inputs of the decoder, where $N$ and $L$ indicate the number of visual embeddings and caption tokens, respectively. The overall network operates according to the following schema: 
\begin{align}
    \text{encoder} \quad \quad & \bm{\tilde{v}}_i = \text{MSA}(\bm{v}_i, \mbb{V}), \nonumber \\
  \addlinespace[0.08cm]
    \text{decoder} \quad \quad & 
    \begin{aligned}
    \mbb{O}_{\bm{w}_i} &= \text{MSCA}(\bm{w}_i, \mbb{\tilde{V}}, \left[ \bm{w}_1, \bm{w}_2, ..., \bm{w}_i \right]), \\
    \end{aligned}
\end{align}
where $\mbb{O}$ is the network output, $\text{MSA}(\bm{x}, \mbb{Y})$ is a self-attention with $\bm{x}$ mapped to query and $\mbb{Y}$ mapped to key-values, $\text{MSCA}(\bm{x}, \mbb{Y}, \mbb{Z})$ indicates a self-attention with $\bm{x}$ as query and $\mbb{Z}$ as key-values, followed by cross-attention with $\bm{x}$ as query and $\mbb{Y}$ as key-values, and $\mbb{\tilde{V}} = \left[ \bm{\tilde{v}}_1, \bm{\tilde{v}}_2, ..., \bm{\tilde{v}}_N\right]$. We omit feed-forward layers and the dependency between consecutive layers for ease of notation.

\begin{figure} [t]
    \centering
    \footnotesize
    \setlength{\tabcolsep}{.2em}
    \resizebox{\linewidth}{!}{
    \begin{tabular}{cccc}
    \textbf{Original Image} & \textbf{Depth Map} & \textbf{Objects} & \textbf{Visual Activation} \\
    \addlinespace[0.05cm]
     \includegraphics[width=0.24\columnwidth]{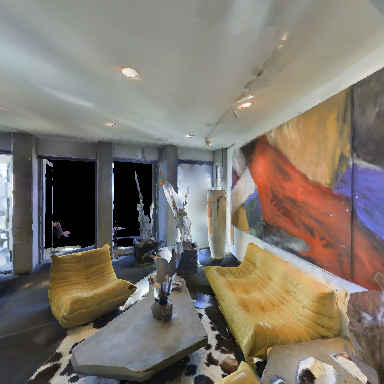} &
     \includegraphics[width=0.24\columnwidth]{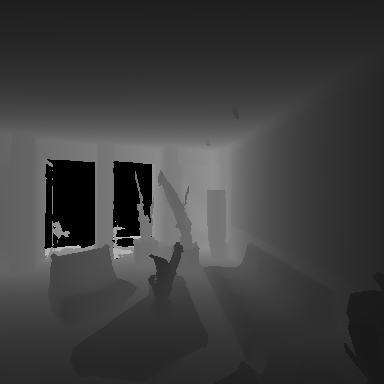} &
     \includegraphics[width=0.24\columnwidth]{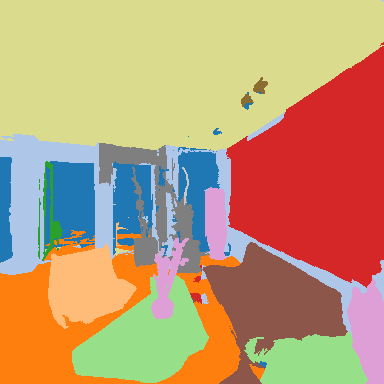} &
     \includegraphics[width=0.24\columnwidth]{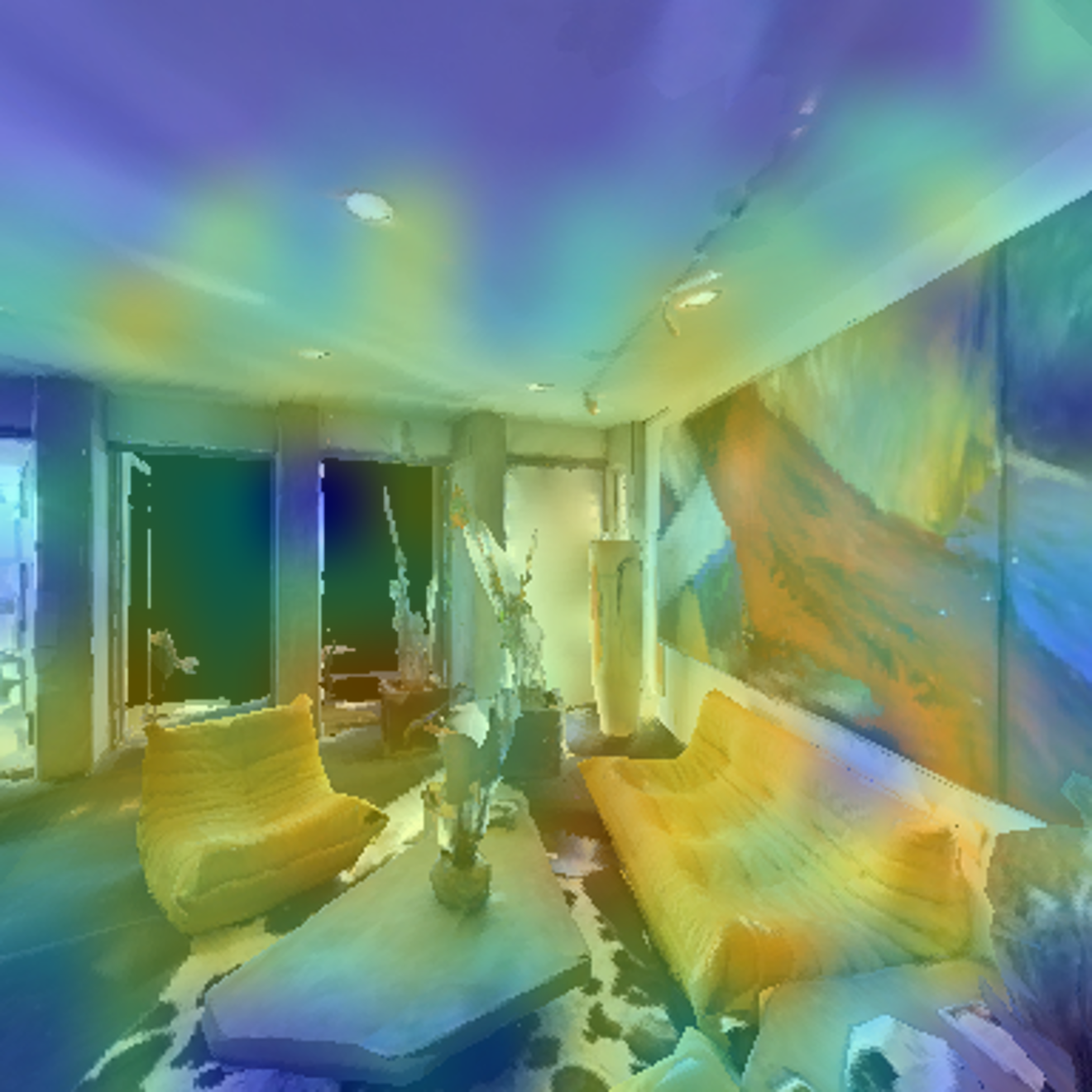} \\
    \end{tabular}
    }
    \caption{A sample of agent observation and corresponding images used by the speaker policy to trigger the captioner.}
    \vspace{-0.5cm}
    \label{fig:samples_policies}
\end{figure}

\subsection{Speaker Policy}
While exploring the environment, the agent sees various RGB observations. Even if the agent was navigating efficiently, the majority of the observations would be overlapped with each other, and the same objects would be observed at multiple consecutive timesteps. Since the agent should describe only relevant scenes during exploration and avoid uninformative captions or unnecessary repetitions, a component that controls caption generation becomes necessary.
We thus introduce a speaker policy which is responsible for triggering the captioner depending on the current view. We compare three approaches that exploit different modalities: a \textit{depth-based} policy, an \textit{object-based} policy, and a \textit{visual activation-based} policy. An example of the considered modalities for the same observation is reported in Fig.~\ref{fig:samples_policies}.

\tit{Depth-based Policy}
High mean depth values indicate a larger area observed by the agent, and potentially, a richer scene to be described. Instead, when the field of view of the agent is occluded by an obstacle, the mean depth value of the observation $s^{d}_t$ is typically low. 
Therefore, the depth-driven policy uses the current depth observation $s^{d}_t$ and computes its mean value. The captioner is activated if the mean depth value is above a predetermined threshold \textbf{\textit{D}}. 

\tit{Object-based Policy}
Considering that the description of the scene will concentrate on relevant objects, the object-driven policy uses the number of relevant objects in the RGB observation $s^{rgb}_t$ to decide if the captioner should generate the description. Specifically, the captioner is triggered only if at least a number \textbf{\textit{O}} of objects are being observed in the scene since using observations with multiple objects allows a larger variety of generated captions.

\tit{Visual Activation-based Policy}
Another possible strategy to implement the speaker policy entails exploiting the activation maps of the same visual encoder used by the captioner (which is a CLIP-like~\cite{radford2021learning} encoder, as detailed in Sec.~\ref{ssec:cap_module}). Such a speaker policy is more closely related to the captioning module and provides a means to interpret the image regions that are more relevant to the agent. In particular, in this work, we consider a CNN-based visual encoder and thus take the feature maps from the last convolutional block, projected into a $d$-dimensional vector. This vector is then averaged, and the speaker policy is activated if its average is above a certain threshold \textbf{\textit{A}}, thus indicating the presence of sufficient semantic content in the image.
\section{Experiments}

\subsection{Implementation and Training Details}

\tinytit{Navigator}
All the exploration models are trained for $5\text{M}$ frames on Gibson Dataset~\cite{xia2018gibson} environments using Habitat simulator~\cite{savva2019habitat}. The evaluation is performed using the test split of Matterport3D (MP3D) dataset~\cite{chang2017matterport3d} and the validation split of Gibson tiny dataset, because they contain object annotations that are used to evaluate the generated captions.

The RGB-D input to the components of the navigator is resized to $128\times128$ pixels, and the global map size is $M=2001$ for MP3D and $M=961$ and Gibson environments. The size of the local map predicted by the mapper is $L=101$, and each pixel in the maps describes a $5 \times 5$ cm\textsuperscript{2} of the environment. Regarding the global policy, the grid size used for the prediction of the global goal is $G=240$, and the global goal is sampled every $N_G=25$ timesteps. 
Both the global and local policies are trained using PPO algorithm~\cite{schulman2017proximal} with a learning rate of $2.5\times10^{-4}$, while the mapper and the pose estimator use a learning rate of $10^{-3}$. Episode length is set to $500$ and $1000$ for the training and evaluation phases, respectively.

\tit{Speaker Policy}
Since the MP3D dataset has richer object annotations and larger environments than the Gibson dataset, we compare two different sets of threshold values depending on the evaluation dataset for the depth- and object-based policies for triggering the captioner. In particular, the threshold values for MP3D dataset are $\textbf{\textit{D}}=(1.0, 2.0, 3.0)$ and $\textbf{\textit{O}}=(1, 3, 5)$ for depth- and object-based policies. Depth and object thresholds are $\textbf{\textit{D}}=(1.0, 1.5, 2.0)$ and $\textbf{\textit{O}}=(1, 2, 3)$ for Gibson dataset. On the other hand, for the activation-based criterion, we use the same set of threshold values for both the evaluation datasets, \ie~$\textbf{\textit{A}}=(4.5, 5.0, 5.5)$.

\tit{Captioner} 
As training and evaluation dataset, we employ COCO~\cite{lin2014microsoft} following the splits defined in~\cite{karpathy2015deep}. To improve the generalization abilities of the model, we also train a variant on a combination of 35.7M images taken from both human-collected datasets (\ie~COCO~\cite{lin2014microsoft}) and web-collected sources (\ie~SBU~\cite{ordonez2011im2text}, CC3M~\cite{sharma2018conceptual}, CC12M~\cite{changpinyo2021conceptual}, WIT~\cite{srinivasan2021wit}, and a subset of YFCC100M~\cite{thomee2016yfcc100m}).

We consider three configurations of the captioner, varying the number of decoding layers $l$, model dimensionality $d$, and the number of attention heads $h$: Tiny ($l=3$, $d=384$, $h=6$), Small ($l=6$, $d=512$, $h=8$), and Base ($l=12$, $d=768$, $h=12$). For all models, we employ CLIP-ViT-L/14~\cite{radford2021learning} as visual feature extractor and three layers in the visual encoder. To assess the effectiveness of CLIP-based features, we also consider a variant of the Tiny model that employs region-based visual features, extracted from Faster R-CNN~\cite{ren2017faster,anderson2018bottom}.
We train all captioning variants with cross-entropy loss using LAMB~\cite{you2019large} as optimizer. We employ the learning rate scheduling strategy proposed in~\cite{vaswani2017attention}, with a warmup of 6,000 iterations and a batch size equal to 1,080. We additionally fine-tune the models with the SCST strategy~\cite{rennie2017self}, by using the Adam optimizer~\cite{kingma2015adam}, a fixed learning rate equal to $5\times 10^{-6}$, and a batch size of 80.

\subsection{Evaluation Protocol}
As the task of smart scene description requires both exploration and description capabilities, for evaluation we use exploration and captioning metrics, as well as a novel score specifically devised for the task.

\tit{Navigator}
As for the performance of the navigation module, we express them in terms of metrics that are commonly used for evaluating embodied exploration agents. In particular, we consider the intersection over union between the ground-truth map of the environment and the map reconstructed by the agent (\textbf{$\mathsf{IoU}$}), the extent of correctly mapped area (\ie~the map accuracy \textbf{$\mathsf{Acc}$}), and the extent of environment area visited by the agent (\ie~the area seen \textbf{$\mathsf{AS}$}), both expressed in $m^2$. 

\tit{Captioner} 
For evaluating the performance of the captioning module on the COCO dataset, we consider the standard image captioning metrics BLEU-4~\cite{papineni2002bleu}, METEOR~\cite{banerjee2005meteor}, ROUGE~\cite{lin2004rouge}, CIDEr~\cite{vedantam2015cider}, and SPICE~\cite{spice2016}.

\tit{Episode Description Score}
Different from standard captioning settings, where ground-truth captions are available for the images, in our setting such information is not available. However, the considered 3D environments datasets come with annotations of the objects in the scene, which can be exploited for performance evaluation. In particular, based on the objects in the scene, we define the \textbf{soft-coverage score} (\textbf{$\mathsf{Cov}$}) and the \textbf{diversity score} (\textbf{$\mathsf{Div}$}), to evaluate the ability of the agent to mention all the relevant objects in the scene and to produce interesting, non-repetitive descriptions, respectively.
The first is computed by considering the intersection score between the set of nouns in the produced caption and the set of categories of the relevant objects in the scene. By ``relevant object'' we mean those whose area covers at least 10\% of the total image area, and thus, can be more useful to identify a scene.
The latter score is defined as the intersection over union between the sets of nouns mentioned in two consecutively generated captions.

Additionally, we measure the agent's overall \textbf{loquacity} (\textbf{$\mathsf{Loq}$}) as the number of times it is activated by the speaker policy, normalized by the episode length. In other words, the loquacity can be seen as the inverse of the average number of navigation steps between two consecutive captions. Moreover, we resort to the recently-proposed \textbf{CLIP score}~\cite{hessel2021clipscore} ($\mathsf{CLIP}\text{-}\mathsf{S}$), in its unpaired definition, to evaluate the alignment between the agent's view and the generated caption.

To evaluate the overall system on each episode, we define an ad-hoc score to measure the concept coverage of the generated descriptions, which is an important aspect of the task.
The proposed \textbf{episode description score} (\textbf{$\mathsf{ED}\text{-}\mathsf{S}$}) reflects the ability of the robot to produce sufficient descriptions in strategic moments, so that the maximum amount of information collected in the environment is covered. The rationale is that it should capture the ability of the agent to mention all the relevant landmarks (objects and rooms) when needed, without unnecessary repetitions. This makes the description more useful and interesting.
The score is defined as:
\begin{equation}
    \mathsf{ED}\text{-}\mathsf{S} = \overline{\mathsf{CLIP}\text{-}\mathsf{S}} \cdot \mathsf{IoU}(\textbf{N}, \textbf{O}) \cdot \text{\%}\mathsf{AS},
\end{equation}
where $\overline{\mathsf{CLIP}\text{-}\mathsf{S}}$ is mean of the CLIP scores of all the captions produced during the episode. Moreover, $\textbf{N}$ is the list of nouns in all the captions produced during the episode and $\textbf{O}$ is the list of objects in the environment. The intersection-over-union operator $\mathsf{IoU}$ is implemented via the Jonker-Volgenant linear assignment algorithm~\cite{jonker1987shortest}. Finally, $ \text{\%}\mathsf{AS}$ is the percentage of the total environment area visited by the agent.
At the dataset level, the $\mathsf{ED}\text{-}\mathsf{S}$ is given by the average of the scores obtained in the dataset episodes.

\begin{table}[t]
\centering
\setlength{\tabcolsep}{.35em}
\resizebox{0.95\linewidth}{!}{
\begin{tabular}{l c ccc c ccc}
\toprule
 & & \multicolumn{3}{c}{\textbf{Gibson Val}} & & \multicolumn{3}{c}{\textbf{MP3D Test}} \\
\cmidrule{3-5} \cmidrule{7-9}
\textbf{Model} & & 
$\mathsf{IoU}$ & $\mathsf{Acc}$ & $\mathsf{AS}$ & & 
$\mathsf{IoU}$ & $\mathsf{Acc}$ & $\mathsf{AS}$ \\
\midrule
Curiosity~\cite{ramakrishnan2021exploration} 
& & 0.528 & 66.19 & 102.59 
& & 0.368 & 130.34 & 186.67\\
Coverage~\cite{chaplot2019learning} 
& & 0.608 & 73.69 & 102.66
& & 0.417 & 146.16 & 195.03 \\
Anticipation~\cite{ramakrishnan2020occupancy} 
& & 0.706 & 81.13 & 102.22
& & 0.494 & 157.02 & 177.14 \\
Impact (Grid)~\cite{bigazzi2022impact} 
& & \textbf{0.738} & \textbf{82.91} & 104.16 
& & \textbf{0.519} & 164.26 & 185.13 \\
Impact (DME)~\cite{bigazzi2022impact} 
& & 0.694 & 79.47 & \textbf{105.03}
& & 0.496 & \textbf{167.58} & \textbf{205.02} \\
\bottomrule
\end{tabular}
}
\caption{Navigation results on Gibson Val and MP3D Test.}
\vspace{-0.2cm}
\label{tab:navigation}
\end{table}

\begin{table}[t]
\centering
\setlength{\tabcolsep}{.25em}
\resizebox{\linewidth}{!}{
\begin{tabular}{lc ccccccc}
\toprule
 & & \makecell{\textbf{Train Ims}} & $\mathsf{BLEU}\text{-}\mathsf{4}$ & $\mathsf{METEOR}$ & $\mathsf{ROUGE}$ & $\mathsf{CIDEr}$ & $\mathsf{SPICE}$ \\
\midrule
Region-based$^\text{tiny}$ & & 112k & 37.7 & 28.3 & 57.6 & 124.8 & 21.9 \\ 
CLIP-based$^\text{tiny}$ & & 112k & 40.6 & 30.0 & 59.9 & 139.4 & 23.9 \\ 
CLIP-based$^\text{small}$ & & 112k & 40.9 & 30.4 & 60.1 & 141.5 & 24.5 \\ 
CLIP-based$^\text{base}$ & & 112k & 41.4 & 30.2 & 60.2 & 142.0 & 24.0 \\ 
CLIP-based$^\text{base}$ & & 35.7M & \textbf{42.9} & \textbf{31.4} & \textbf{61.5} & \textbf{149.6} & \textbf{25.0} \\
\bottomrule
\end{tabular}
}
\caption{Captioning results on the COCO test set.}
\label{tab:captioning}
\vspace{-0.5cm}
\end{table}

\subsection{Experimental Results}

\tit{Navigation Results}
First, we compare the different exploration approaches alone on the MP3D and Gibson datasets. The results of this analysis are reported in Table~\ref{tab:navigation}.
The best agent in terms of the area seen (\textbf{$\mathsf{AS}$}) is the impact-based method using density model estimation. In particular, this approach is able to efficiently explore both Gibson and MP3D datasets, giving its best in large environments. In fact, the small $0.87 \text{m}^2$ margin over the second best approach on Gibson becomes $9.99 \text{m}^2$ in the larger MP3D environments. Moreover, this method is still competitive in terms of \textbf{$\mathsf{IoU}$}, being also the best in terms of \textbf{$\mathsf{Acc}$} on the MP3D test split. In light of these results, we use the impact-based navigator with DME as the navigator of the overall approach.

\tit{Captioning Results}
Then, we evaluate the performance of the captioner alone on the COCO dataset. The results of this analysis are reported in Table~\ref{tab:captioning}. It can be observed that the CLIP-based variants are the best-performing ones, with a noticeable advantage over the region-based captioner. This confirms the representative power of CLIP features. The Base variant has also been trained on additional image-caption pairs from web-collected sources, which further increases its performance.
It is worth mentioning that these results are in line with those of state-of-the-art captioners (\eg~\cite{li2020oscar,zhang2021vinvl}).
In light of these results, we use the CLIP-based Base variant as the captioner of the overall approach.

\begin{table}[t]
\setlength{\tabcolsep}{.3em}
\resizebox{\linewidth}{!}{
\begin{tabular}{lcc cccc c cccc}
\toprule
& & & \multicolumn{4}{c}{\textbf{COCO only}} & & \multicolumn{4}{c}{\textbf{COCO + Web-collected}} \\
\cmidrule{3-7} \cmidrule{9-12}
 & $\mathsf{Loq}$ & & $\mathsf{Cov}$ & $\mathsf{Div}$ & $\mathsf{CLIP}\text{-}\mathsf{S}$ & $\mathsf{ED}\text{-}\mathsf{S}$  & & $\mathsf{Cov}$ & $\mathsf{Div}$ & $\mathsf{CLIP}\text{-}\mathsf{S}$ & $\mathsf{ED}\text{-}\mathsf{S}$ \\
\midrule
\textbf{Always} & 100.00 & & 0.864 & 0.352 & 0.670 & 0.119 & & 0.862 & 0.348 & 0.692 & 0.120 \\
\midrule
\textbf{Depth} \\
\hspace{0.3cm}$\mathbf{D\geq1.0}$ & 83.26 & & 0.868 & 0.335 & 0.670 & 0.140 & & 0.865 & 0.338 & 0.690 & 0.140 \\
\hspace{0.3cm}$\mathbf{D\geq1.5}$ & 55.24 & & 0.871 & 0.323 & 0.664 & 0.203 & & 0.868 & 0.330 & 0.683 & 0.204 \\
\hspace{0.3cm}$\mathbf{D\geq2.0}$ & 27.38 & & 0.793 & 0.293 & 0.629 & 0.250 & & 0.780 & 0.304 & 0.650 & 0.257 \\
\midrule
\textbf{Object} \\
\hspace{0.3cm}$\mathbf{O\geq1}$ & 41.73 & & 0.793 & 0.314 & 0.663 & 0.222 & & 0.784 & 0.332 & 0.682 & 0.225 \\
\hspace{0.3cm}$\mathbf{O\geq2}$ & 21.55 & & 0.703 & 0.289 & 0.645 & 0.219 & & 0.697 & 0.307 & 0.664 & 0.220 \\
\hspace{0.3cm}$\mathbf{O\geq3}$ &  7.58 & & 0.416 & 0.232 & 0.549 & 0.107 & & 0.410 & 0.260 & 0.561 & 0.105 \\
\midrule
\textbf{Activation} \\
\hspace{0.3cm}$\mathbf{A\geq4.5}$ & 87.79 & & 0.866 & 0.340 & 0.672 & 0.134 & & 0.864 & 0.343 & 0.691 & 0.134 \\
\hspace{0.3cm}$\mathbf{A\geq5.0}$ & 51.13 & & 0.828 & 0.349 & 0.674 & 0.223 & & 0.827 & 0.348 & 0.691 & 0.220 \\
\hspace{0.3cm}$\mathbf{A\geq5.5}$ &  2.20 & & 0.133 & 0.153 & 0.455 & 0.038 & & 0.140 & 0.153 & 0.464 & 0.040 \\
\bottomrule
\end{tabular}
}
\centering
\caption{Episode description results on Gibson tiny validation set.} 
\label{tab:ex2_results_gibson}
\vspace{-0.15cm}
\end{table}

\begin{table}[t]
\setlength{\tabcolsep}{.3em}
\resizebox{\linewidth}{!}{
\begin{tabular}{lcc cccc c cccc}
\toprule
& & & \multicolumn{4}{c}{\textbf{COCO only}} & & \multicolumn{4}{c}{\textbf{COCO + Web-collected}} \\
\cmidrule{3-7} \cmidrule{9-12}
 & $\mathsf{Loq}$ & & $\mathsf{Cov}$ & $\mathsf{Div}$ & $\mathsf{CLIP}\text{-}\mathsf{S}$ & $\mathsf{ED}\text{-}\mathsf{S}$ &
 & $\mathsf{Cov}$ & $\mathsf{Div}$ & $\mathsf{CLIP}\text{-}\mathsf{S}$ & $\mathsf{ED}\text{-}\mathsf{S}$  \\
\midrule
 \textbf{Always} & 100.00 & & 0.768 & 0.363 & 0.648 & 0.172 & & 0.771 & 0.348 & 0.687 & 0.179 \\
\midrule
\textbf{Depth} \\
\hspace{0.3cm}$\mathbf{D\geq1.0}$ & 89.05 & & 0.765 & 0.352 & 0.648 & 0.180 & &
                                              0.767 & 0.341 & 0.687 & 0.187 \\
\hspace{0.3cm}$\mathbf{D\geq2.0}$ & 45.06 & & 0.751 & 0.317 & 0.637 & 0.155 & & 
                                              0.750 & 0.311 & 0.668 & 0.160 \\
\hspace{0.3cm}$\mathbf{D\geq3.0}$ & 15.98 & & 0.317 & 0.161 & 0.338 & 0.030 & & 
                                              0.317 & 0.151 & 0.360 & 0.031 \\
\midrule
\textbf{Object} \\
\hspace{0.3cm}$\mathbf{O\geq1}$ & 75.82 & & 0.754 & 0.340 & 0.635 & 0.190 & &
                                            0.756 & 0.333 & 0.670 & 0.196 \\
\hspace{0.3cm}$\mathbf{O\geq3}$ & 46.57 & & 0.700 & 0.310 & 0.605 & 0.168 & & 
                                            0.701 & 0.310 & 0.634 & 0.172 \\
\hspace{0.3cm}$\mathbf{O\geq5}$ & 19.90 & & 0.616 & 0.255 & 0.533 & 0.106 & & 
                                            0.614 & 0.254 & 0.553 & 0.107 \\
\midrule
\textbf{Activation} \\
\hspace{0.3cm}$\mathbf{A\geq4.5}$ & 82.05 & & 0.765 & 0.348 & 0.641 & 0.106 & & 
                                              0.767 & 0.337 & 0.676 & 0.107 \\
\hspace{0.3cm}$\mathbf{A\geq5.0}$ & 46.28 & & 0.754 & 0.350 & 0.636 & 0.153 & & 
                                              0.757 & 0.341 & 0.667 & 0.158 \\
\hspace{0.3cm}$\mathbf{A\geq5.5}$ &  1.28 & & 0.325 & 0.118 & 0.347 & 0.015 & & 
                                              0.328 & 0.116 & 0.362 & 0.016 \\
\bottomrule
\end{tabular}
}
\centering
\caption{Episode description results on MP3D test set.} 
\vspace{-0.5cm}
\label{tab:ex2_results_mp3d}
\end{table}

\begin{figure} [t]
    \centering
    \includegraphics[width=0.965\columnwidth]{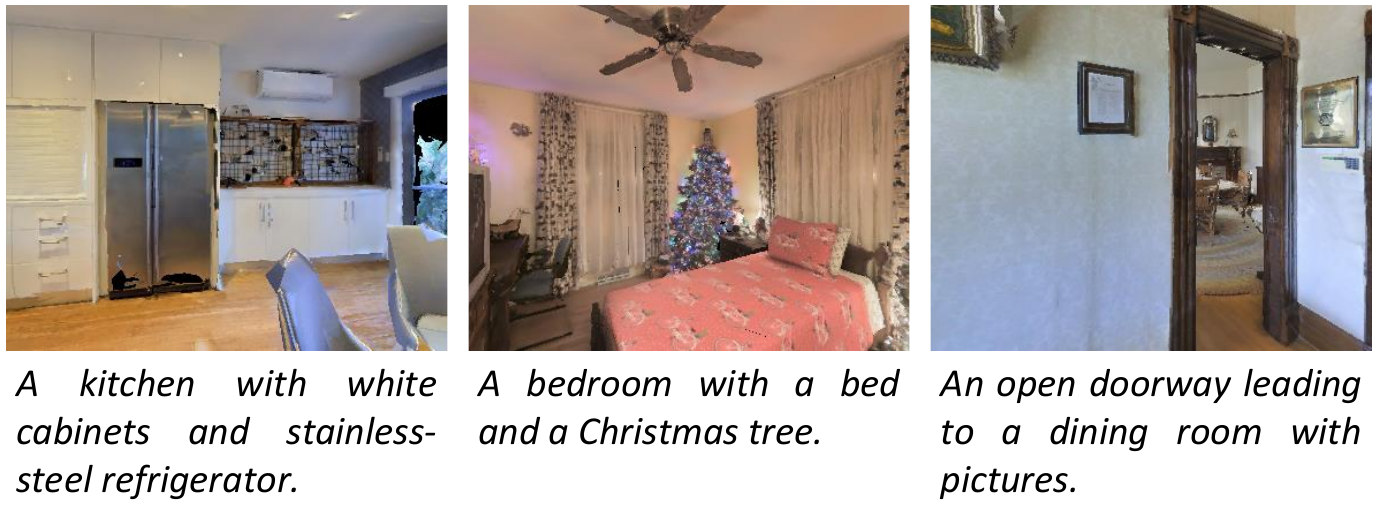}
    \caption{Sample observations and corresponding captions generated by our model.}
    \vspace{-0.5cm}
    \label{fig:exemplar_captions}
\end{figure}

\tit{Episode Description Results} Finally, we compare variants of the overall approach using different speaking policies with different threshold values, and use as reference a dummy policy according to which the captioning module is always activated. The results are reported in Tables~\ref{tab:ex2_results_gibson} and~\ref{tab:ex2_results_mp3d}. It can be noticed that the captioner trained on web-collected sources performs better that the variant trained on COCO only in terms of all metrics, suggesting its superior generalization capabilities and thus, suitability to be employed in an embodied setting. However, to evaluate on the overall task, the proposed $\mathsf{ED}\text{-}\mathsf{S}$ score is more informative than the other metrics, which can nonetheless be used in combination with the $\mathsf{ED}\text{-}\mathsf{S}$ to gain additional insights on the agents' behaviour. In fact, the values of all metrics but the $\mathsf{ED}\text{-}\mathsf{S}$ are comparable in both datasets, while the $\mathsf{ED}\text{-}\mathsf{S}$ is on average higher on Gibson: this is due to the fact that Gibson has on average smaller and less cluttered spaces, which can be more easily fully explored (higher values of the $\mathsf{Cov}$ on Gibson confirm this intuition). This trend is further confirmed by the fact that on the Gibson dataset the speaking policy must ensure the $\mathsf{Loc}$ being in a specific range (roughly between 20 and 80) to obtain the best $\mathsf{ED}\text{-}\mathsf{S}$ scores, while on the wider spaces of MP3D, speaking policies ensuring a higher $\mathsf{Loc}$ lead to better performance. Qualitative examples of the output of our approach on selected observations are reported in Fig.~\ref{fig:exemplar_captions}.

\tit{Real-World Deployment}
Exploration agents trained on the photorealistic environments of the Habitat simulator and general-purpose captioners allow the deployment of our approach to the real world, using a LoCobot platform~\cite{locobot}. For the deployment, the captioner is left untouched, whilst we modify the camera parameters of the navigator, such as camera height, the field of view, and depth sensor range to match the real-world setting. Furthermore, the deployed agent is trained by adding noise models fitted to mimic the LoCobot camera noise over the observations retrieved from the simulator. As the last step, we apply the correction presented by~\cite{bigazzi2021out} to correct noisy real-world depth observations. In the video accompanying the submission, we show the agent exploring and describing a real-world apartment.
\section{Conclusion}
\label{sec:conclusion}
In this work, we have presented an embodied exploration agent whose internal representation of the environment can be interpreted by non-expert users. This is achieved by equipping the agent with the ability to produce a natural language description of the observed scene when this is deemed interesting according to a speaking policy. The experimental results show that the proposed approach is a viable solution to gain insights into the perception and navigation capabilities of embodied agents. Moreover, the generalization capabilities of the modules adopted allow real-world deployment without major redesigns.


\clearpage
\bibliographystyle{IEEEtran}
\bibliography{bibliography.bib}

\begin{thebibliography}{10}
\providecommand{\url}[1]{#1}
\csname url@rmstyle\endcsname
\providecommand{\newblock}{\relax}
\providecommand{\bibinfo}[2]{#2}
\providecommand\BIBentrySTDinterwordspacing{\spaceskip=0pt\relax}
\providecommand\BIBentryALTinterwordstretchfactor{4}
\providecommand\BIBentryALTinterwordspacing{\spaceskip=\fontdimen2\font plus
\BIBentryALTinterwordstretchfactor\fontdimen3\font minus
  \fontdimen4\font\relax}
\providecommand\BIBforeignlanguage[2]{{%
\expandafter\ifx\csname l@#1\endcsname\relax
\typeout{** WARNING: IEEEtran.bst: No hyphenation pattern has been}%
\typeout{** loaded for the language `#1'. Using the pattern for}%
\typeout{** the default language instead.}%
\else
\language=\csname l@#1\endcsname
\fi
#2}}

\bibitem{xia2018gibson}
F.~Xia, A.~R. Zamir, Z.~He, A.~Sax, J.~Malik, and S.~Savarese, ``{Gibson Env:
  Real-world perception for embodied agents},'' in \emph{Proceedings of the
  IEEE/CVF Conference on Computer Vision and Pattern Recognition}, 2018.

\bibitem{chang2017matterport3d}
A.~Chang, A.~Dai, T.~Funkhouser, M.~Halber, M.~Niessner, M.~Savva, S.~Song,
  A.~Zeng, and Y.~Zhang, ``{Matterport3D: Learning from RGB-D Data in Indoor
  Environments},'' in \emph{Proceedings of the International Conference on 3D
  Vision}, 2017.

\bibitem{savva2019habitat}
M.~Savva, A.~Kadian, O.~Maksymets, Y.~Zhao, E.~Wijmans, B.~Jain, J.~Straub,
  J.~Liu, V.~Koltun, J.~Malik, \emph{et~al.}, ``{Habitat: A Platform for
  Embodied AI Research},'' in \emph{Proceedings of the IEEE/CVF International
  Conference on Computer Vision}, 2019.

\bibitem{wijmans2019dd}
E.~Wijmans, A.~Kadian, A.~Morcos, S.~Lee, I.~Essa, D.~Parikh, M.~Savva, and
  D.~Batra, ``{DD-PPO: Learning Near-Perfect PointGoal Navigators from 2.5
  Billion Frames},'' in \emph{Proceedings of the International Conference on
  Learning Representations}, 2019.

\bibitem{chaplot2019learning}
D.~S. Chaplot, D.~Gandhi, S.~Gupta, A.~Gupta, and R.~Salakhutdinov, ``{Learning
  To Explore Using Active Neural SLAM},'' in \emph{Proceedings of the
  International Conference on Learning Representations}, 2019.

\bibitem{ramakrishnan2020occupancy}
S.~K. Ramakrishnan, Z.~Al-Halah, and K.~Grauman, ``{Occupancy Anticipation for
  Efficient Exploration and Navigation},'' in \emph{Proceedings of the European
  Conference on Computer Vision}, 2020.

\bibitem{ramakrishnan2021exploration}
S.~K. Ramakrishnan, D.~Jayaraman, and K.~Grauman, ``{An Exploration of Embodied
  Visual Exploration},'' \emph{International Journal of Computer Vision}, vol.
  129, no.~5, pp. 1616--1649, 2021.

\bibitem{bigazzi2022impact}
R.~Bigazzi, F.~Landi, S.~Cascianelli, L.~Baraldi, M.~Cornia, and R.~Cucchiara,
  ``{Focus on Impact: Indoor Exploration with Intrinsic Motivation},''
  \emph{IEEE Robotics and Automation Letters}, vol.~7, no.~2, pp. 2985--2992,
  2022.

\bibitem{kadian2020sim2real}
A.~Kadian, J.~Truong, A.~Gokaslan, A.~Clegg, E.~Wijmans, S.~Lee, M.~Savva,
  S.~Chernova, and D.~Batra, ``{Sim2Real Predictivity: Does evaluation in
  simulation predict real-world performance?}'' \emph{IEEE Robotics and
  Automation Letters}, vol.~5, no.~4, pp. 6670--6677, 2020.

\bibitem{bigazzi2021out}
R.~Bigazzi, F.~Landi, M.~Cornia, S.~Cascianelli, L.~Baraldi, and R.~Cucchiara,
  ``{Out of the Box: Embodied Navigation in the Real World},'' in
  \emph{Proceedings of the International Conference on Computer Analysis of
  Images and Patterns}, 2021.

\bibitem{truong2021bi}
J.~Truong, S.~Chernova, and D.~Batra, ``Bi-directional domain adaptation for
  sim2real transfer of embodied navigation agents,'' \emph{IEEE Robotics and
  Automation Letters}, vol.~6, no.~2, pp. 2634--2641, 2021.

\bibitem{anderson2018vision}
P.~Anderson, Q.~Wu, D.~Teney, J.~Bruce, M.~Johnson, N.~S{\"u}nderhauf, I.~Reid,
  S.~Gould, and A.~van~den Hengel, ``Vision-and-language navigation:
  Interpreting visually-grounded navigation instructions in real
  environments,'' in \emph{Proceedings of the IEEE/CVF Conference on Computer
  Vision and Pattern Recognition}, 2018.

\bibitem{landi2019perceive}
F.~Landi, L.~Baraldi, M.~Cornia, M.~Corsini, and R.~Cucchiara, ``{Multimodal
  Attention Networks for Low-Level Vision-and-Language Navigation},''
  \emph{Computer Vision and Image Understanding}, vol. 210, p. 103255, 2021.

\bibitem{krantz2020beyond}
J.~Krantz, E.~Wijmans, A.~Majumdar, D.~Batra, and S.~Lee, ``{Beyond the
  Nav-Graph: Vision-and-Language Navigation in Continuous Environments},'' in
  \emph{Proceedings of the European Conference on Computer Vision}, 2020.

\bibitem{anderson2021sim}
P.~Anderson, A.~Shrivastava, J.~Truong, A.~Majumdar, D.~Parikh, D.~Batra, and
  S.~Lee, ``{Sim-to-Real Transfer for Vision-and-Language Navigation},'' in
  \emph{Proceedings of the Conference on Robot Learning}, 2021.

\bibitem{das2018embodied}
A.~Das, S.~Datta, G.~Gkioxari, S.~Lee, D.~Parikh, and D.~Batra, ``{Embodied
  Question Answering},'' in \emph{Proceedings of the IEEE/CVF Conference on
  Computer Vision and Pattern Recognition Workshops}, 2018.

\bibitem{das2018neural}
A.~Das, G.~Gkioxari, S.~Lee, D.~Parikh, and D.~Batra, ``{Neural Modular Control
  for Embodied Question Answering},'' in \emph{Proceedings of the Conference on
  Robot Learning}, 2018.

\bibitem{wijmans2019embodied}
E.~Wijmans, S.~Datta, O.~Maksymets, A.~Das, G.~Gkioxari, S.~Lee, I.~Essa,
  D.~Parikh, and D.~Batra, ``{Embodied Question Answering in Photorealistic
  Environments With Point Cloud Perception},'' in \emph{Proceedings of the
  IEEE/CVF Conference on Computer Vision and Pattern Recognition}, 2019.

\bibitem{yu2019multi}
L.~Yu, X.~Chen, G.~Gkioxari, M.~Bansal, T.~L. Berg, and D.~Batra,
  ``{Multi-Target Embodied Question Answering},'' in \emph{Proceedings of the
  IEEE/CVF Conference on Computer Vision and Pattern Recognition}, 2019.

\bibitem{cascianelli2018full}
S.~Cascianelli, G.~Costante, T.~A. Ciarfuglia, P.~Valigi, and M.~L. Fravolini,
  ``{Full-GRU Natural Language Video Description for Service Robotics
  Applications},'' \emph{IEEE Robotics and Automation Letters}, vol.~3, no.~2,
  pp. 841--848, 2018.

\bibitem{cornia2019smart}
M.~Cornia, L.~Baraldi, and R.~Cucchiara, ``{SMArT: Training Shallow
  Memory-aware Transformers for Robotic Explainability},'' in \emph{Proceedings
  of the IEEE International Conference on Robotics and Automation}, 2020.

\bibitem{bigazzi2020explore}
R.~Bigazzi, F.~Landi, M.~Cornia, S.~Cascianelli, L.~Baraldi, and R.~Cucchiara,
  ``{Explore and Explain: Self-supervised Navigation and Recounting},'' in
  \emph{Proceedings of the International Conference on Pattern Recognition},
  2020.

\bibitem{tan2022embodied}
S.~Tan, D.~Guo, H.~Liu, X.~Zhang, and F.~Sun, ``Embodied scene description,''
  \emph{Autonomous Robots}, vol.~46, pp. 21--43, 2022.

\bibitem{karpathy2015deep}
A.~Karpathy and L.~Fei-Fei, ``Deep visual-semantic alignments for generating
  image descriptions,'' in \emph{Proceedings of the IEEE/CVF Conference on
  Computer Vision and Pattern Recognition}, 2015.

\bibitem{xu2015show}
K.~Xu, J.~Ba, R.~Kiros, K.~Cho, A.~Courville, R.~Salakhutdinov, R.~S. Zemel,
  and Y.~Bengio, ``Show, attend and tell: Neural image caption generation with
  visual attention,'' in \emph{Proceedings of the International Conference on
  Machine Learning}, 2015.

\bibitem{anderson2018bottom}
P.~Anderson, X.~He, C.~Buehler, D.~Teney, M.~Johnson, S.~Gould, and L.~Zhang,
  ``{Bottom-up and top-down attention for image captioning and visual question
  answering},'' in \emph{Proceedings of the IEEE/CVF Conference on Computer
  Vision and Pattern Recognition}, 2018.

\bibitem{vaswani2017attention}
A.~Vaswani, N.~Shazeer, N.~Parmar, J.~Uszkoreit, L.~Jones, A.~N. Gomez,
  {\L}.~Kaiser, and I.~Polosukhin, ``Attention is all you need,'' in
  \emph{Advances in Neural Information Processing Systems}, 2017.

\bibitem{liu2021cptr}
W.~Liu, S.~Chen, L.~Guo, X.~Zhu, and J.~Liu, ``{CPTR: Full Transformer Network
  for Image Captioning},'' \emph{arXiv preprint arXiv:2101.10804}, 2021.

\bibitem{cornia2021universal}
M.~Cornia, L.~Baraldi, G.~Fiameni, and R.~Cucchiara, ``{Universal Captioner:
  Inducing Content-Style Separation in Vision-and-Language Model Training},''
  \emph{arXiv preprint arXiv:2111.12727}, 2022.

\bibitem{rennie2017self}
S.~J. Rennie, E.~Marcheret, Y.~Mroueh, J.~Ross, and V.~Goel, ``Self-critical
  sequence training for image captioning,'' in \emph{Proceedings of the
  IEEE/CVF Conference on Computer Vision and Pattern Recognition}, 2017.

\bibitem{huang2019attention}
L.~Huang, W.~Wang, J.~Chen, and X.-Y. Wei, ``{Attention on Attention for Image
  Captioning},'' in \emph{Proceedings of the IEEE/CVF International Conference
  on Computer Vision}, 2019.

\bibitem{cornia2020meshed}
M.~Cornia, M.~Stefanini, L.~Baraldi, and R.~Cucchiara, ``{Meshed-Memory
  Transformer for Image Captioning},'' in \emph{Proceedings of the IEEE/CVF
  Conference on Computer Vision and Pattern Recognition}, 2020.

\bibitem{zhang2021rstnet}
X.~Zhang, X.~Sun, Y.~Luo, J.~Ji, Y.~Zhou, Y.~Wu, F.~Huang, and R.~Ji,
  ``{RSTNet: Captioning with Adaptive Attention on Visual and Non-Visual
  Words},'' in \emph{Proceedings of the IEEE/CVF Conference on Computer Vision
  and Pattern Recognition}, 2021.

\bibitem{ramakrishnan2022poni}
S.~K. Ramakrishnan, D.~S. Chaplot, Z.~Al-Halah, J.~Malik, and K.~Grauman,
  ``{PONI: Potential Functions for ObjectGoal Navigation with Interaction-free
  Learning},'' in \emph{Proceedings of the IEEE/CVF Conference on Computer
  Vision and Pattern Recognition}, 2022.

\bibitem{he2016deep}
K.~He, X.~Zhang, S.~Ren, and J.~Sun, ``Deep residual learning for image
  recognition,'' in \emph{Proceedings of the IEEE/CVF Conference on Computer
  Vision and Pattern Recognition}, 2016.

\bibitem{ronneberger2015u}
O.~Ronneberger, P.~Fischer, and T.~Brox, ``{U-Net: Convolutional Networks for
  Biomedical Image Segmentation},'' in \emph{Proceedings of the International
  Conference on Medical Image Computing and Computer Assisted Intervention},
  2015.

\bibitem{pathak2017curiosity}
D.~Pathak, P.~Agrawal, A.~A. Efros, and T.~Darrell, ``Curiosity-driven
  exploration by self-supervised prediction,'' in \emph{Proceedings of the
  International Conference on Machine Learning}, 2017.

\bibitem{chaplot2020object}
D.~S. Chaplot, D.~P. Gandhi, A.~Gupta, and R.~R. Salakhutdinov, ``{Object Goal
  Navigation using Goal-Oriented Semantic Exploration},'' in \emph{Advances in
  Neural Information Processing Systems}, 2020.

\bibitem{raileanu2020ride}
R.~Raileanu and T.~Rockt{\"a}schel, ``{RIDE: Rewarding impact-driven
  exploration for procedurally-generated environments},'' in \emph{Proceedings
  of the International Conference on Learning Representations}, 2021.

\bibitem{radford2021learning}
A.~Radford, J.~W. Kim, C.~Hallacy, A.~Ramesh, G.~Goh, S.~Agarwal, G.~Sastry,
  A.~Askell, P.~Mishkin, J.~Clark, G.~Krueger, and I.~Sutskever, ``{Learning
  Transferable Visual Models From Natural Language Supervision},'' in
  \emph{Proceedings of the International Conference on Machine Learning}, 2021.

\bibitem{zhang2021vinvl}
P.~Zhang, X.~Li, X.~Hu, J.~Yang, L.~Zhang, L.~Wang, Y.~Choi, and J.~Gao,
  ``{VinVL: Revisiting visual representations in vision-language models},'' in
  \emph{Proceedings of the IEEE/CVF Conference on Computer Vision and Pattern
  Recognition}, 2021.

\bibitem{schulman2017proximal}
J.~Schulman, F.~Wolski, P.~Dhariwal, A.~Radford, and O.~Klimov, ``{Proximal
  Policy Optimization Algorithms},'' \emph{arXiv preprint arXiv:1707.06347},
  2017.

\bibitem{lin2014microsoft}
T.-Y. Lin, M.~Maire, S.~Belongie, J.~Hays, P.~Perona, D.~Ramanan,
  P.~Doll{\'a}r, and C.~L. Zitnick, ``{Microsoft COCO: Common Objects in
  Context},'' in \emph{Proceedings of the European Conference on Computer
  Vision}, 2014.

\bibitem{ordonez2011im2text}
V.~Ordonez, G.~Kulkarni, and T.~Berg, ``{Im2Text: Describing Images Using 1
  Million Captioned Photographs},'' in \emph{Advances in Neural Information
  Processing Systems}, 2011.

\bibitem{sharma2018conceptual}
P.~Sharma, N.~Ding, S.~Goodman, and R.~Soricut, ``{Conceptual Captions: A
  Cleaned, Hypernymed, Image Alt-text Dataset For Automatic Image
  Captioning},'' in \emph{Proceedings of the Annual Meeting of the Association
  for Computational Linguistics}, 2018.

\bibitem{changpinyo2021conceptual}
S.~Changpinyo, P.~Sharma, N.~Ding, and R.~Soricut, ``{Conceptual 12M: Pushing
  Web-Scale Image-Text Pre-Training To Recognize Long-Tail Visual Concepts},''
  in \emph{Proceedings of the IEEE/CVF Conference on Computer Vision and
  Pattern Recognition}, 2021.

\bibitem{srinivasan2021wit}
K.~Srinivasan, K.~Raman, J.~Chen, M.~Bendersky, and M.~Najork, ``{WIT:
  Wikipedia-based Image Text Dataset for Multimodal Multilingual Machine
  Learning},'' in \emph{Proceedings of the International ACM SIGIR Conference
  on Research and Development in Information Retrieval}, 2021.

\bibitem{thomee2016yfcc100m}
B.~Thomee, D.~A. Shamma, G.~Friedland, B.~Elizalde, K.~Ni, D.~Poland, D.~Borth,
  and L.-J. Li, ``{YFCC100M: The new data in multimedia research},''
  \emph{Communications of the ACM}, vol.~59, no.~2, pp. 64--73, 2016.

\bibitem{ren2017faster}
S.~Ren, K.~He, R.~Girshick, and J.~Sun, ``{Faster R-CNN: towards real-time
  object detection with region proposal networks},'' \emph{IEEE Transactions on
  Pattern Analysis and Machine Intelligence}, vol.~39, no.~6, pp. 1137--1149,
  2017.

\bibitem{you2019large}
Y.~You, J.~Li, S.~Reddi, J.~Hseu, S.~Kumar, S.~Bhojanapalli, X.~Song,
  J.~Demmel, K.~Keutzer, and C.-J. Hsieh, ``{Large Batch Optimization for Deep
  Learning: Training BERT in 76 minutes},'' in \emph{Proceedings of the
  International Conference on Learning Representations}, 2019.

\bibitem{kingma2015adam}
D.~Kingma and J.~Ba, ``Adam: a method for stochastic optimization,'' in
  \emph{Proceedings of the International Conference on Learning
  Representations}, 2015.

\bibitem{papineni2002bleu}
K.~Papineni, S.~Roukos, T.~Ward, and W.-J. Zhu, ``{BLEU: a method for automatic
  evaluation of machine translation},'' in \emph{Proceedings of the Annual
  Meeting of the Association for Computational Linguistics}, 2002.

\bibitem{banerjee2005meteor}
S.~Banerjee and A.~Lavie, ``{METEOR: An automatic metric for MT evaluation with
  improved correlation with human judgments},'' in \emph{Proceedings of the
  Annual Meeting of the Association for Computational Linguistics Workshops},
  2005.

\bibitem{lin2004rouge}
C.-Y. Lin, ``Rouge: A package for automatic evaluation of summaries,'' in
  \emph{Proceedings of the Annual Meeting of the Association for Computational
  Linguistics Workshops}, 2004.

\bibitem{vedantam2015cider}
R.~Vedantam, C.~Lawrence~Zitnick, and D.~Parikh, ``{CIDEr: Consensus-based
  Image Description Evaluation},'' in \emph{Proceedings of the IEEE/CVF
  Conference on Computer Vision and Pattern Recognition}, 2015.

\bibitem{spice2016}
P.~Anderson, B.~Fernando, M.~Johnson, and S.~Gould, ``{SPICE: Semantic
  Propositional Image Caption Evaluation},'' in \emph{Proceedings of the
  European Conference on Computer Vision}, 2016.

\bibitem{hessel2021clipscore}
J.~Hessel, A.~Holtzman, M.~Forbes, R.~L. Bras, and Y.~Choi, ``{CLIPScore: A
  Reference-free Evaluation Metric for Image Captioning},'' in
  \emph{Proceedings of the Conference on Empirical Methods in Natural Language
  Processing}, 2021.

\bibitem{jonker1987shortest}
R.~Jonker and A.~Volgenant, ``A shortest augmenting path algorithm for dense
  and sparse linear assignment problems,'' \emph{Computing}, vol.~38, no.~4,
  pp. 325--340, 1987.

\bibitem{li2020oscar}
X.~Li, X.~Yin, C.~Li, P.~Zhang, X.~Hu, L.~Zhang, L.~Wang, H.~Hu, L.~Dong,
  F.~Wei, \emph{et~al.}, ``{Oscar: Object-semantics aligned pre-training for
  vision-language tasks},'' in \emph{Proceedings of the European Conference on
  Computer Vision}, 2020.

\bibitem{locobot}
``{LoCoBot: An Open Source Low Cost Robot},''
  \url{https://locobot-website.netlify.com}.

\end{thebibliography}

\end{document}